\documentclass[final]{cvpr}
\usepackage{comment}
\usepackage{times}
\usepackage{epsfig}
\usepackage{graphicx}
\usepackage{amsmath}
\usepackage{amssymb}

\usepackage{enumitem}

\usepackage[pagebackref=true,breaklinks=true,colorlinks,bookmarks=false]{hyperref}

\def\cvprPaperID{****} 
\def\confYear{CVPR 2021}

\begin{document}

\title{
Exploring Data Pipelines through the Process Lens: a Reference Model for Computer Vision}

\author{Agathe Balayn\\
TU Delft\\
{\tt\small a.m.a.balayn@tudelft.nl}
\and
Bogdan Kulynych\\
EPFL\\
{\tt\small bogdan.kulynych@epfl.ch}
\and
Seda G{\"u}rses\\
TU Delft\\
{\tt\small F.S.Gurses@tudelft.nl}
}

\maketitle

\begin{abstract}
Researchers have identified datasets used for training computer vision (CV) models as an important source of hazardous outcomes, and continue to examine popular CV datasets to expose their harms. These works tend to treat datasets as objects, or focus on particular steps in data production pipelines. We argue here that we could further systematize our analysis of harms by examining CV data pipelines through a process-oriented lens that captures the creation, the evolution and use of these datasets. 
As a step towards cultivating a process-oriented lens, we embarked on an empirical study of CV data pipelines informed by the field of method engineering. We present here a preliminary result: a reference model of CV data pipelines. Besides exploring the questions that this endeavor raises, we discuss how the process lens could support researchers in discovering understudied issues, and could help practitioners in making their processes more transparent.
\end{abstract}

\section{Introduction}\label{sec:intro}
Training data can have a considerable impact on the outputs of a machine learning system, by affecting its accuracy, or causing undesirable or even harmful effects when the system is deployed in applications that act in the world. 

In response to these harms, machine-learning  
and interdisciplinary researchers who tackle the societal impact of computer vision  typically a) explore characteristics of
datasets with respect to their potentially harmful impacts on society~\cite{crawford2019excavating,shankar2017no,maleve2020data}, e.g., taxonomies with offensive labels, often resulting from \emph{uncritical practices}~\cite{birhane2021large}; b) identify and mitigate harms by adapting single, discrete processes in the data pipeline, e.g. filtering out the offensive and non-visual labels of the taxonomy, 
~\cite{yang2020towards,birhane2021large,jo2020lessons}; c) propose techniques to make datasets more transparent~\cite{gebru2018datasheets},  to encourage analysts to be more accountable and reflective of potential harms.

These forms of enquiry presuppose considering 
\textit{datasets as objects} with configurable and conceivable properties. 
Recent exceptions \cite{milagros2021documenting, sambasivan2021everyone,crawford2018anatomy,denton2020bringing,hutchinson2020towards} hint at another direction, that explores not only the datasets themselves but considers the \emph{data pipelines} and actors behind their creation and use more broadly.
In this paper, we take inspiration from these works and explore a \emph{process-oriented lens} on data pipelines. 
Such a lens may encourage researchers and practitioners to be more rigorous in the production and the analysis of datasets vis a vis their potential societal impact.
Most importantly, a process lens will enable 
to recognize and capture the following:

\textit{Datasets as complex, living objects.}
Datasets are not inert but living objects. They are regularly updated in production to improve models, while they are often re-purposed and adapted for new applications.  
Moreover, datasets interact with each other, particularly through models, e.g. they are composed using transfer learning~\cite{li2020transfer} such as for improving facial recognition systems meant to be applied in a wide variety of contexts~\cite{ren2014transfer}.
However, while datasets can create harms at each phase of their life, academic works on datasets typically are ambiguous about these phases and seem to be based only on their first release.

\textit{The pipeline as a vector of harms.}
Not only the processes of the pipeline that follow the first release might raise issues, but also the  interplay between these processes and the ones leading to the first release, as 
it defines the potential errors of a model. 
For instance, in Detroit, following harmful mistakes of a facial recognition system in deployment
, the police has decided to only apply it to still images, as it is closer to the training data collected in a static setting in development
, which shows how some harms can occur due to problematic mismatches between processes in deployment and development phases~\cite{nprfacial}. 
Some existing works have already investigated the impact of pipeline processes
such as rotation, cropping, and other pre-processing on models' accuracy
\cite{engstrom2019exploring,hendrycks2018benchmarking,zheng2016improving}, although not necessarily on societal harms~\cite{crawford2019excavating}.  
Nonetheless, little attention was given to studying the impact of sequences of processes. 

\textit{Reasoning behind data pipeline design.}
The lack of transparency about the design of a dataset's pipeline (see examples in \autoref{sec:sketches}) makes it challenging for researchers to analyse the procedural harms or downstream harms of models training on a dataset.
Such transparency would reveal not only the processes themselves, but also the reasoning behind them.
For instance, while it is valuable to study datasets in-depth, it can be intractable to analyse each label and image due to the scale.
Documentation of the reasoning behind the choices of label ontologies, and the process of collecting corresponding images, would improve the analysis. E.g., knowing that WordNet was used to select the labels of ImageNet~\cite{deng2009imagenet} enables us to directly reflect on harms from the point of view of the taxonomy~\cite{crawford2019excavating}, and 
knowing that images were collected from country-specific search engines enables us to foresee the cultural skews in label representations~\cite{shankar2017no}. Such matters are 
not consistently included in academic works, and they are not always explicitly asked for in documentation methodologies~\cite{gebru2018datasheets}.



\smallskip

\textbf{Our contributions.}
In this paper, we devise a systematic approach to study the data pipelines in computer vision. For this, we combine techniques from \emph{the development of reference models}, \emph{method engineering},  and \emph{process engineering}.
These are  well-established, intertwined fields~\cite{winter2006reference} that  propose methods to capture complex sets of processes. 
Using these techniques, we
develop an empirically informed reference model of data pipelines in computer vision.
We then reflect on its potential usefulness, as well as the possible limitations of this approach for creating transparency and accountability for practitioners and researchers.


\section{Methodology}
Our objective is to build a reference model of the data pipeline in machine learning-based computer vision tasks. The literature on method, process and reference-model engineering identifies the following requirements towards reference models:
\emph{representativity} of the pipeline used in practice at an abstraction level high enough for including \emph{comprehensively} each of their processes~\cite{mackenzie2006reference,gholami2010procedure,brinkkemper1996method}, 
\emph{modularity} to easily add new processes~\cite{henderson2010situational,brinkkemper1996method},  
\emph{clarity}  
for individuals interested in different parts of the pipeline~\cite{mackenzie2006reference,brinkkemper1996method},
and sufficient level of \emph{detail} to enable actionable reporting, and 
reflection on potential ethical issues. 


Following these requirements, we build:  a high-level \emph{reference model} that serves as an ontology of our main concepts
; a set of low-level models of each identifiable process within existing pipelines; a set of \emph{process maps} reflecting the pipelines of the main computer vision practices. 

\smallskip
\textbf{Populating the models and maps.}
The methods from each of the mentioned fields have some advantages and limitations.
\emph{Process patterns} in process engineering allow for a fine-grained description of the processes, but generally impose an order on these processes, which does not correspond to the computer vision pipelines we analysed. 
\emph{Reference models} do not necessarily impose an order, but can be too generic for our desired level of detail. 
However, the \emph{method engineering} formalisms can alleviate these issues as they support processes of different granularities, and the interactions between the processes can be represented in a \emph{map}  
without necessarily enforcing order.

Thus, we employed a mix of methods from the different disciplines. The details 
can be found in \autoref{sec:method}. In short:
1) We did a systematic literature review that presents computer vision datasets (51 publications) or specific processes to build datasets (16 publications) selected out of 220 relevant papers found using different search methods. 
2) We identified and noted the main processes within each publication. 
3) By comparing each description, we iteratively identified common processes, 
refined their granularity, and aggregated them into meaningful method processes. 
4) We described each 
of them 
with a relevant formalism, and drew the maps of each dataset pipeline. 
5) We searched for grey areas and are in the process of conducting expert interviews to further validate and refine our reference model, and to model processes that are not objects of scientific publications (e.g. processes for serving data in deployment). 

\section{Results: Reference model of data processes}\label{sec:reference_model}

\begin{figure*}
    \centering
    \includegraphics[width=1\textwidth]{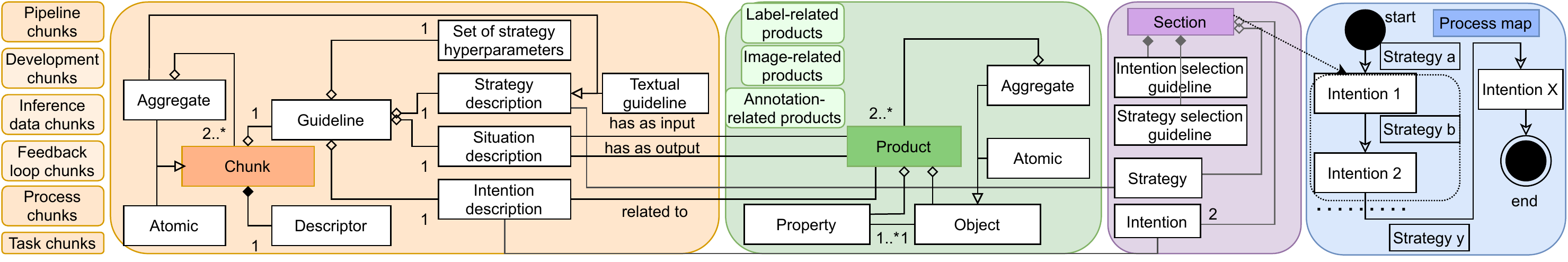}
    \caption{UML diagram describing the main components of our reference model formalism, and activity diagram for the process map. Relationships: \rule[0.5ex]{1em}{0.55pt} association, $\lozenge$ aggregation (i.e. ``part-of''), $\blacklozenge$ composition,  $\vartriangleright$ inheritance (i.e. ``is-a'').} 
    \label{fig:UML}
\end{figure*}

In this section, we present our formalism of the components of the reference model summarized in \autoref{fig:UML}, and provide examples of its application.

\subsection{Products}
The \emph{products} are the essential components of the datasets. Each product is defined by an \emph{object} (
e.g. an image, a set of images, an image-annotation pair, etc.) and a set of \emph{properties}. Based on the literature review, we have identified the following types of products:

\textit{Label-related products.}
A dataset comprises a {set of target labels}, often organized into a {hierarchical taxonomy}, with higher levels of the taxonomy representing more abstract concepts (e.g. animal) and lower levels representing more fine-grained concepts (e.g. types of animals).
The main properties considered in the reviewed literature are the size ({number of labels}) and the {number of levels} in the hierarchy.
Many properties of labels are currently implicit, 
whereas other properties, like where the taxonomy originates from, can be relevant for capturing potential harms. 

\textit{Image-related products.}
A dataset typically contains a {{set of images}}. The {size of the set}, the {content of the images} and especially its diversity are  put forward as main properties in the reviewed literature. 
Additional properties can be considered, especially the {metadata} attached to the images (e.g. author of the images, etc.), {physical characteristics of the images} such as their {quality, resolution}, etc., and the {distribution of these characteristics} across the set. 

\textit{Annotation-related products.}
These products can be thought as a {list of tuples, each composed of an image and one or multiple annotations}. The properties put forward are often the {number and quality of annotations per image}.


\subsection{Chunks} 
Borrowing a standard term from method engineering, we refer to the processes of the data pipeline as \emph{chunks}. They can be composed: a chunk can consist of multiple lower-level chunks.
We define 3 major \emph{granularity levels}. 
1) The pipeline level---the main sections of the data pipeline. These are the development of training data, 
the processes of improving and updating the training data in deployment (\emph{feedback loops}), and the processing of data at inference time. 
2) The process level---the individual activities taking place in the chunks of the pipeline level (e.g., data collection).
3) The task level---the sub-activities required to conduct the activities of the prior level (e.g. query preparation and image crawling for data collection). These chunks can be further divided into sub-tasks when necessary.

Each chunk has an associated \emph{guideline} that defines the activity it represents in relation to the products; and a \emph{descriptor} that uniquely identifies it. 
A guideline is composed of: i) an \emph{intention description}, i.e. the goal of the chunk, ii)  a \emph{situation description}, i.e. the input and output products of the chunk,
iii) a \emph{strategy description} that   describes the activity textually (i.e. on the finest granularity), or indicates the composition of a sequence of lower level chunks to perform the activity, iv) a set of \emph{strategy hyperparameters} allowing to be precise about the design choices in the strategy. 

Based on the literature review, we have identified the following process-level chunks: label definition, data collection, data annotation, data filtering, data processing, data augmentation, data splitting, and product refinement. We provide a short description of these, and some task-level chunks that we have identified in \autoref{sec:activity}.



\subsection{Process maps}
Each dataset is created in a pipeline that uses different chunks, in different orders, at different granularities. Composing chunks in various ways
impacts the final dataset products and the outputs of a resulting model. We formalise the overall process, i.e. the connections between chunks, through \emph{intention-strategy maps} and their \emph{sections}, as inspired from method engineering that assembles chunks to compose individual situated methods. We develop these maps to  materialize the process lens mentioned in the \autoref{sec:intro}, and to evaluate its potential and limitations.


Intuitively, an intention-strategy map represents a sequence of chunks. 
Formally, the intention-strategy map is an ordered sequence of intentions connected by the strategies to realize them (see \autoref{fig:UML}), each intention-strategy pair referring to a chunk guideline. 
Attached to the sequence are \emph{sections}. Each section consists of the initial and  subsequent \emph{intention} of each of the chunks, and the \emph{strategy} to realize the subsequent intention, i.e. the description of how the chunk fulfills its intention. Additionally, the section also  contains an \emph{intention selection guideline} specifying the reason to go from the first intention to the second, and a \emph{strategy selection guideline} specifying the reason for choosing this particular strategy to perform this intention.

\begin{figure*}[h!]
    \centering
    \includegraphics[width=0.9\textwidth]{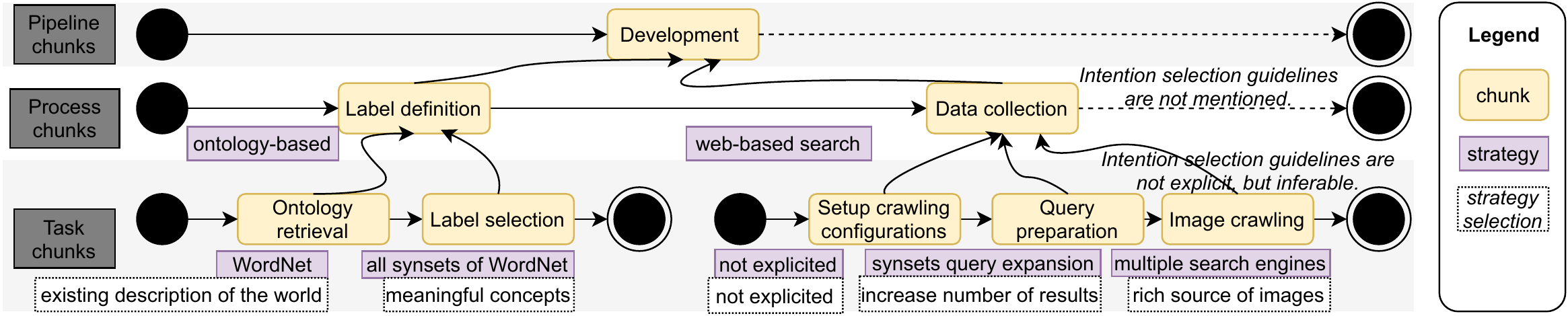}
    \caption{Example segments of the process map associated to the development of ImageNet~\cite{deng2009imagenet}.}
    \label{fig:imagenet}
\end{figure*}

As an illustration, \autoref{fig:imagenet} provides an example of a segment of the process map for the ImageNet creation process.

\smallskip

We identified multiple chunks and process maps in our literature review, which shows that a plurality of sequences exists. This suggests that there is no standardized data pipeline in computer vision. The process-level chunks are not always the same, and the order might differ widely or may not be reported at all.
The data-processing chunks, for example, happen either after collection, after annotation, or after filtering. The task-level chunks and their parameters may also vary between the training and deployment phases. While both may impact the outputs of the model, they are often not clearly specified in the papers.


\section{Discussion}
Introducing processes via the formalism presents advantages  for researchers and practitioners of computer vision.

\subsection{Insights for researchers}

\textit{Identification of knowledge gaps.}
Our formalism enables to systematically analyze typical processes employed in computer vision across tasks. This unifying lens uncovers the grey areas that escape scrutiny. In fact, we believe that only a fraction of existing 
process chunks have been analyzed on the subject of ethical considerations and downstream negative consequences~\cite{paullada2020data}. 
For example, our framework highlighted an understudied issue: 
the papers we reviewed rarely report on data processing chunks---yet the interaction of these chunks with other pipeline chunks can potentially have a significant impact. For instance, cropping images after labeling might exclude the visual information relevant to the label, making the model learn wrong associations, a matter that seems to not have been studied previously. Our formalism allows to surface  this issue as it makes apparent the related chunks
, and the current lack of reporting on them. 

\textit{Process details.}
Applying our formalism also uncovers what chunks, metrics, and intentions the researchers deem worth documenting in the publications. This can guide researchers towards identifying the chunks that should merit more detailed descriptions. For example, while a few interdisciplinary publications mention issues with the label-related products, less than 10\% of the reviewed papers were found to refer to some of the properties of these products, e.g. only the SUN database~\cite{xiao2010sun} and MS-COCO~\cite{lin2014microsoft} outline their reasoning around  the completeness of the set of labels employed. 
As for data collection and filtering, only the authors of Tiny Images~\cite{torralba200880} make transparent the potential biases in images retrieved from search engines, and  include a study of the dataset noise with and without filtering these images. 
A more rigorous documentation could support a more systematic analysis of potential issues.

\subsection{Insights for practitioners}
Supporting developers or auditors in making processes surrounding data more transparent and structured with our formalism could help with capturing or fostering reflection around potential issues.
Compared to previous frameworks~\cite{gebru2018datasheets}, our formalism enables to report on the entire pipeline surrounding the dataset instead of solely the dataset  or a subset of  processes executed to develop it. 
For instance, it outlines both the development and deployment pipelines, which makes it easier to identify potential differences in terms of chunk, chunk order, and chunk hyperparameters between them. These differences can lead to distribution shift, resulting in misclassifications, unfairness in outputs, or non-robustness, as the Detroit police example showed.

The body of documented chunks of an organization can grow over time. The identical format and reusable nature of chunks may help practitioners to easily identify the chunks of a new pipeline, and to build on existing documentation. It may also encourage them to share ethical considerations corresponding to chunks across pipelines.



\subsection{Reflections and limitations}
Despite potential advantages, the feasibility of building a satisfying 
reference model for systematizing the analysis of data pipelines for harms 
requires further discussion.  
Most papers we studied, and our preliminary interviews with practitioners, have shown 
a divergence in data pipeline design, including lack of common chunks or order of processes. 
This has parallels in software engineering. Prior work in method engineering recognizes that all projects are different and cannot be supported by a single method, but instead 
adapted methodological guidance should be proposed~\cite{brinkkemper1996method}.
However, the divergences in CV data pipelines pose challenges in fulfilling our requirements towards representativity, comprehensiveness, and clarity.
This raises a number of questions. For example: Can we develop a reference model that would capture all these messy or idiosyncratic processes? Would its use in production be tedious and easily out of date? Could modularity help in incorporating further processes and details? Would such a model give sufficient structure to the reporting of the data pipelines for purposes of increased transparency and accountability? 



Finally, our primary interest lies in exploring the use of such formalisms in  systematizing the identification of harms that may stem from data pipelines and surrounding processes. Whether such use of a reference model would aid in more effective and efficient identification and action on harms in data pipelines requires further study.
Other fields such as information retrieval have developed
a common language and benchmarks to study their systems of interest and foster communication between industry, academia and governments, enabling progress on their respective problems~\cite{trec}.
Whether this would also apply to computer vision and analysis of associated harms
is yet to be seen.

Our empirical analysis of data pipelines in this paper has illustrated some of the envisioned benefits of a process-oriented lens. The sole fact that the analysis has also raised the diversity of questions above hints at the richness of this lens, that we plan to investigate in the future.

\newpage

{\small
\bibliographystyle{ieee_fullname}
\bibliography{bibli}
}

\appendix
\section{The Art of Sketching Data Pipelines}\label{sec:sketches}

\begin{figure}[h]
    \centering
    
    \includegraphics[width=\linewidth]{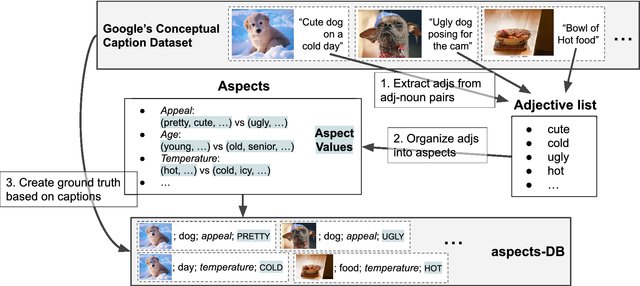}
    \caption{Dataset creation pipeline from \cite{blandfort2020focus}.}
    \label{fig:im_publi}
\end{figure}

\begin{figure}[h]
    \centering
    
    \includegraphics[width=\linewidth]{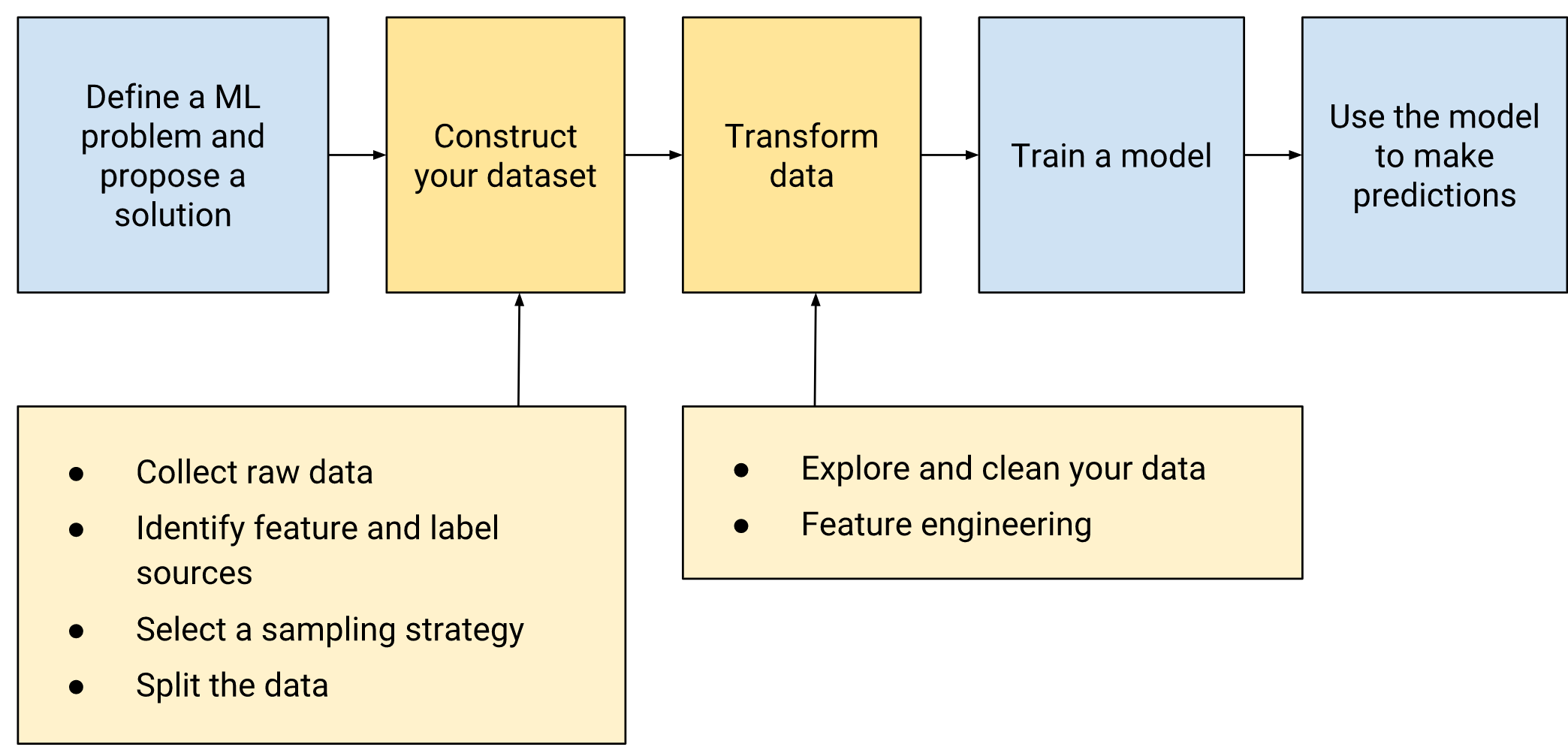}
    \caption{Dataset creation pipeline from a Google course on machine learning {\url{https://developers.google.com/machine-learning/data-prep/process?hl=fr}} }
    \label{fig:google}
\end{figure}

\begin{figure}[h]
    \centering
    
    \includegraphics[width=\linewidth]{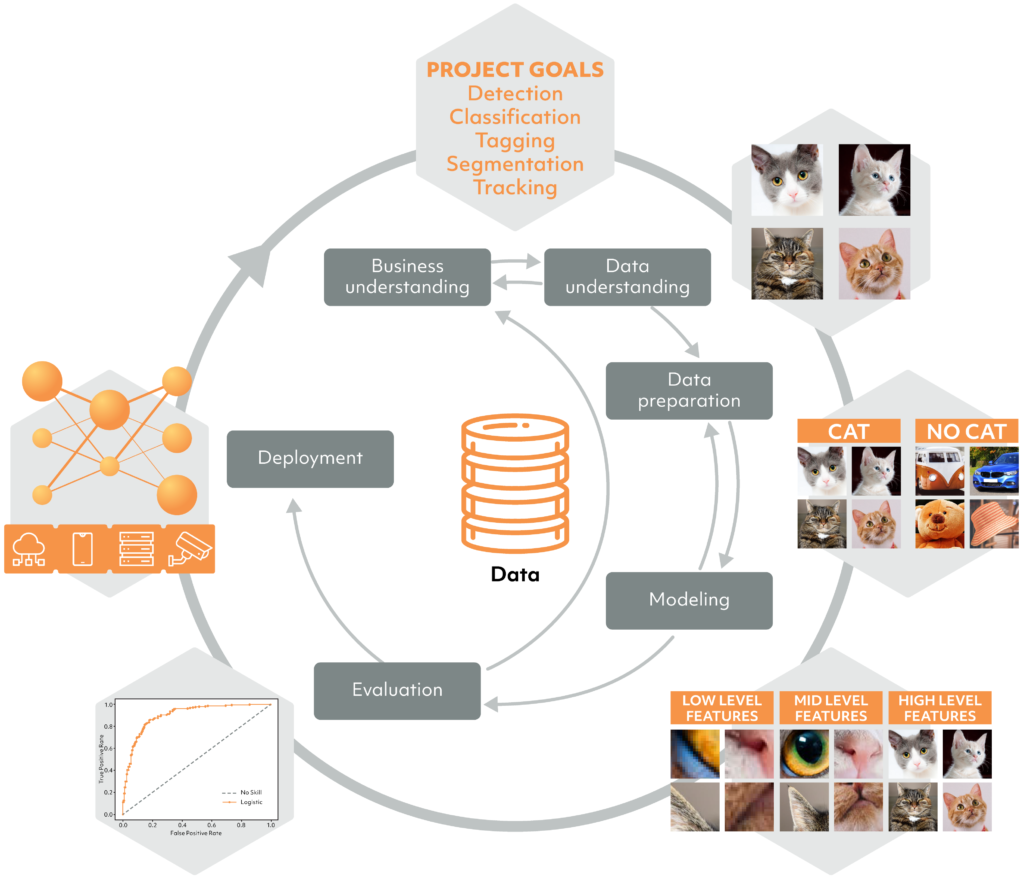}
    \caption{Dataset creation pipeline from Dynam.AI {\url{https://www.dynam.ai/computer-vision-projects-management-part-1/}}, B2B company that implements computer vision solutions.}
    \label{fig:dynamai}
\end{figure}

\begin{figure*}[h]
    \centering
    
    \includegraphics[width=0.7\linewidth]{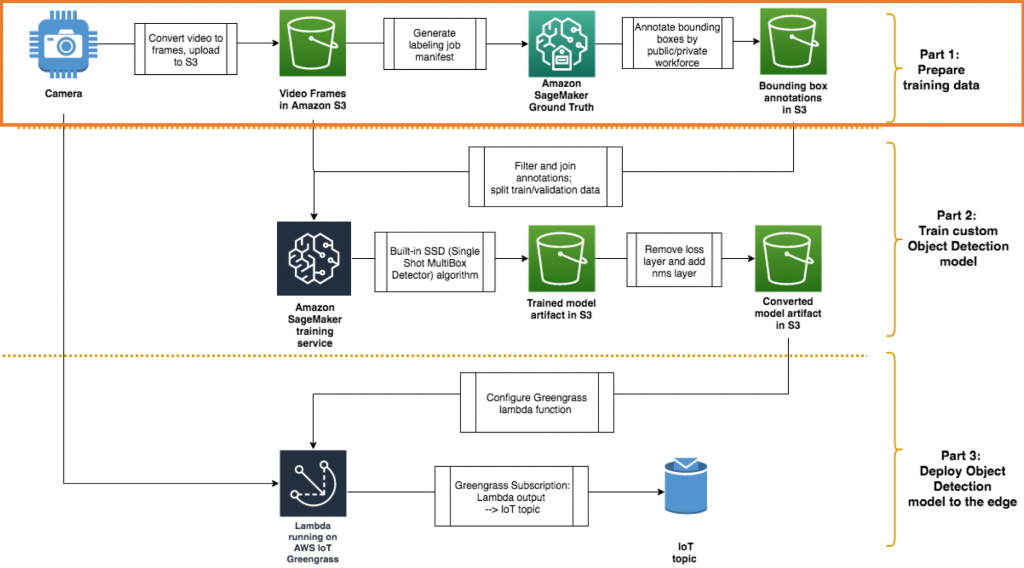}
    \caption{Dataset creation pipeline from Amazon {\url{https://aws.amazon.com/blogs/iot/sagemaker-object-detection-greengrass-part-1-of-3/}} }
    \label{fig:amazon}
\end{figure*}

It is difficult to even find  sketches of data pipelines in publications~\autoref{fig:im_publi}. They are easier to find in grey materials of companies, such as for Amazon~\autoref{fig:amazon}, Google~\autoref{fig:google}, and DynamAI~\autoref{fig:dynamai}. We show examples of these sketches below.

The main observations one can make out of these sketches is that they are very diverse, without any uniformity. They do not all use the same level of granularity to talk about the data pipelines, they do not all use the same vocabulary,  and they refer to various different processes, often mentioning the training data development process but not the deployment processes.

This variability in the sketches does not allow to reflect on the potential harms of the pipelines in a global way, that can be re-used to talk about different pipelines.



\section{Details about the methodology}\label{sec:method}
\subsection{Systematic review of the literature}
 We systematically surveyed the computer science publications that present computer vision datasets, or that propose methods pertaining to the creation of datasets.
 
For that, we went through the list of publications of the following top conferences CVPR, ECCV, ICCV, NeurIPS, ICLR and associated workshops in the general field of machine learning and in the field of computer vision more specifically --the assumption here is that in the more general conferences, some computer vision publications might appear, and some papers might mention general data pipelines  with example in computer vision--, for the years 2015 to June 2020. 
Then we proceeded to snowball sampling from the references of the papers initially collected. 

We only retained publications related to images (we excluded video dataset papers and 3D based works), that touch upon classification or segmentation tasks, as the data  pipelines for other types of data and tasks vary even more than just for these image tasks. It amounted to around 270  papers.
We filtered out other types of tasks such as referring expression comprehension~\cite{chen2020cops}, image captioning and visual question answering because the data  pipelines of many of these datasets are not described in details in the papers, and are not always published in the field of computer vision but also natural language processing. 
The set of applications of this pool of publications (around 220 papers) is large. We further scoped the study by selecting datasets and research publications that have as object of study objects (things), scenes and stuffs, or persons (mainly faces and emotions), as they are one of the main  current interests in computer vision research. 
We filtered out datasets for autonomous driving and aerial views as they consist mostly of road and field views, which is a more restricted set of classification labels and hence present mostly a subset of processes retrieved from the publications we selected (and there are --too-- many of them), and a subset of potential harms. 
Although works relying on these types of data and / or application are not included and we assume that many points we learn from our survey will apply to them, it would surely be beneficial to repeat our analysis on these datasets in the future, as even more harms might arise from them.

In the end, we selected 51 dataset papers to study thoroughly. We also analysed shorter datasets of natural topics such as animals (e.g.~\cite{khan2020animalweb}) and flora~\cite{thapafgvc}, as we noticed that although they serve for the same type of classification tasks, their collection process differs from the other works. 

Concerning computer vision papers that relate to processes of the pipelines, we solely found papers that studied and proposed ways to optimize the crowdsourcing components of the pipelines amounting to the number of 16. Only one paper~\cite{buckler2017reconfiguring} seems to study the impact of the image signal processing pipeline on the accuracy and energy consumption of the following learning pipeline.

\subsection{Interviews}
As for the interviews, we plan to interview two groups of individuals: researchers working in the field of computer vision, and practitioners who develop computer vision pipelines.
In both cases, our goal is to identify limitations of the initial reference model and its population that we drew from the literature review.
For that, we first ask the participants to talk about the pipelines they know of, and to describe them (this way, we can also identify the processes they deem worth reporting).
We then present to the interview participants a drawing of the reference model, and we ask them to discuss it critically, i.e. to identify chunks that could be missing or hyperparameters that could be different, as well as to re-order the chunks based on their experience.
This way, we can iteratively improve our reference model and validate it.
Finally, we interrogate them around harms they know of in relation to the different processes.

\section{Reference model -further details}\label{sec:activity}

\begin{itemize}[leftmargin=*,topsep=0pt, noitemsep]
    \item {Label definition.} 
This chunk consists in defining the label-related products.
The {method employed to define such objects} (which maps to various sub-chunks depending on the dataset), their {source}, and the {individuals who define them} are the main hyperparameters.

\item{Data collection.} 
It consists in collecting a set of potential image-related products for the dataset.
A multitude of hyperparameters lay within this activity, stemming again from {the source of images, the individuals collecting them}, and the general {approach to gather them}.

\item{Data annotation.} 
It consists in collecting the set of image-annotation tuples of the dataset. It is not needed when images are photographed by the researchers for pre-defined labels, but it is when they are collected from the Internet. This chunk strategy is instantiated in different task-chunks, with  an automatic process (pre-trained machine learning models, or meta data of the images), a human process or a hybrid one. 
The human process chunks can then be defined into sub-chunks.

\item{Data filtering.} It consists in fine-tuning the set of image products by removing images that do not fit the requirements fixed for the dataset. It impacts the properties of the image-related  and possibly annotation-related products. 

\item{Data processing.} It consists in transforming the individual images in the image-related products according to certain {processing criteria} for the images to verify certain properties usually in terms of size (downsampling or upsampling), and possibly color hue, lighting, etc. This is done to support their use within machine learning classifiers, and in some cases to simplify the learning tasks.

\item{Data augmentation.} Although  not an activity mentioned in dataset papers, it is a standard practice in industry, and re-uses task-chunks of data processing for modifying image-related products (especially the size of the set).

\item{Data splitting.} The dataset is divided  in a training(, validation) and test sets, that create new versions of all the dataset products. 

\item{Product refinement.}
Although not considered in any computer vision paper, there is an additional set of chunks related to dataset transformations 
generally happening after deployment. Datasets might be reduced in size, augmented with new labels and images, or deprived from some, as ethical or performance issues are identified or distribution shifts happen. 
This impacts the properties of each dataset product. 
Surfacing these chunks might provide insights to develop best practices or critical views on the handling of harms.


\end{itemize}

\end{document}